\documentclass[conference]{IEEEtran}
\IEEEoverridecommandlockouts
\usepackage{cite}
\usepackage{amsmath,amssymb,amsfonts}
\usepackage{algorithmic}
\usepackage{graphicx}
\usepackage[hidelinks]{hyperref}       
\usepackage{varioref}
\usepackage[capitalise]{cleveref}
\usepackage{float}          
\usepackage{subcaption}
\usepackage{booktabs}       
\usepackage{textcomp}
\usepackage{xcolor}
\usepackage{pgfplots}
\usepackage{flushend}
\usepackage{tikz}
\def\BibTeX{{\rm B\kern-.05em{\sc i\kern-.025em b}\kern-.08em
    T\kern-.1667em\lower.7ex\hbox{E}\kern-.125emX}}
\usepackage[toc,section=section]{glossaries}
\usepackage{comment}   

\input{macros.tex}

\makenoidxglossaries 
\newacronym{nfc}{NFC}{Nozzle Failure Classification}
\newacronym{nf1}{NF1}{Nozzle Failure 1}
\newacronym{nf2}{NF2}{Nozzle Failure 2}
\newacronym{nf3}{NF3}{Nozzle Failure 3}
\newacronym{nf4}{NF4}{Nozzle Failure 4}
\newacronym{nf5}{NF5}{Nozzle Failure 5}

\newacronym{paa}{PAA}{Piecewise Aggregate Approximation}
\newacronym{sax}{SAX}{Symbolic Aggregate approXimation}
\newacronym{tsc}{TSC}{Time-Series Classification}
\newacronym{pf}{PF}{Proximity Forest}

\newacronym{dtw}{DTW}{Dynamic Time Warping}
\newacronym{rbf}{RBF}{Radial Basis Function}
\newacronym{svm}{SVM}{Support Vector Machine}
\newacronym{rf}{RF}{Random Forest}
\newacronym{lr}{LR}{Logistic Regression}
\newacronym{dt}{DT}{Decision Tree}
\newacronym{et}{ET}{Extremely Randomized Trees}
\newacronym{knn}{KNN}{K-Nearest Neighbors}
\newacronym{ovr}{OVR}{One-vs-Rest}
\newacronym{ml}{ML}{Machine Learning}
\newacronym{rnn}{RNN}{Reccurrent Neural Network}
\newacronym{lstm}{LSTM}{Long Short-Term Memory}

\newacronym{cpp}{CPP}{Canon Production Printing}
\newacronym{qpv}{QP{\&}V}{Quality Processes {\&} Validation}
\newacronym{eda}{EDA}{Exploratory Data Analysis}
\newacronym{doa}{DOA}{Dead On Arrival}
\newacronym{eol}{EoL}{End-of-Life}

\newacronym{sam}{SAM}{Spectral Angle Mapper}
\newacronym{ann}{ANN}{Artifical Neural Network}
\newacronym{dl}{DL}{Deep Learning}
\newacronym{pca}{PCA}{Principal Component Analysis}
\newacronym{lda}{LDA}{Linear Discriminant Analysis}

\newacronym{p}{P}{Precision}
\newacronym{r}{R}{Recall}
\newacronym{tp}{TP}{True Positive}
\newacronym{fn}{FN}{False Negative}
\newacronym{fp}{FP}{False Positive}
\newacronym{tn}{TN}{True Negative}
\newacronym{auc}{AUC}{Area Under the ROC Curve}
\newacronym{loocv}{LOOCV}{leave-one-out cross validation}

\floatstyle{plaintop}
\restylefloat{table}

\begin{document}

\title{Machine Learning for Pattern Detection in Printhead Nozzle Logging}

\author{
\IEEEauthorblockN{Nikola Prianikov, Marcin Pietrasik, Charalampos S. Kouzinopoulos}
\IEEEauthorblockA{\textit{Department of Advanced Computing Sciences} \\
\textit{Maastricht University}\\
Maastricht, The Netherlands }
\and
\IEEEauthorblockN{Evelyne Janssen}
\IEEEauthorblockA{\textit{Quality Processes and Validation Department} \\
\textit{Canon Production Printing}\\
Venlo, The Netherlands }}

\maketitle

\author{\IEEEauthorblockN{Nikola Prianikov\IEEEauthorrefmark{1},
Evelyne Janssen\IEEEauthorrefmark{2}, Marcin Pietrasik\IEEEauthorrefmark{1},
Charalampos S. Kouzinopoulos\IEEEauthorrefmark{1}}
\IEEEauthorblockA{\IEEEauthorrefmark{1}Department of Advanced Computing Sciences, Maastricht University, Maastricht, The Netherlands}
\IEEEauthorblockB{\IEEEauthorrefmark{2}Quality Processes and Validation Department, Canon Production Printing, Venlo, The Netherlands}}

\maketitle

\begin{abstract}
Correct identification of failure mechanisms is essential for manufacturers to ensure the quality of their products. Certain failures of printheads developed by Canon Production Printing can be identified from the behavior of individual nozzles, the states of which are constantly recorded and can form distinct patterns in terms of the number of failed nozzles over time, and in space in the nozzle grid. In our work, we investigate the problem of printhead failure classification based on a multifaceted dataset of nozzle logging and propose a Machine Learning classification approach for this problem. We follow the feature-based framework of time-series classification, where a set of time-based and spatial features was selected with the guidance of domain experts. Several traditional ML classifiers were evaluated, and the One-vs-Rest Random Forest was found to have the best performance. The proposed model outperformed an in-house rule-based baseline in terms of a weighted F1 score for several failure mechanisms.
\end{abstract}

\begin{IEEEkeywords}
feature engineering, time-series classification, corrective maintenance, printing, nozzle log
\end{IEEEkeywords}

\section{Introduction}

Identifying failure mechanisms is a critical part of industrial corrective maintenance for manufacturers to ensure the quality of their products \cite{er_ratby_impact_2025}. Many manufactured systems are made up of smaller components that can fail over time, eventually causing entire system breakdown. By monitoring the condition of these individual parts, unique patterns of wear and tear can be uncovered, which can then be linked to potential causes of failure. An example of this is the inkjet printheads produced by \gls*{cpp} for high-volume printing. The performance of these printheads depends heavily on the individual drop-forming nozzles, whose failure logging is constantly recorded. By analyzing nozzle failure development patterns in time and space over the nozzle grid from the logging data, failure mechanisms can be identified. The \gls*{cpp} team has observed several distinct patterns of nozzle failures that correspond to the failure mechanism of a complete printhead. For instance, one pattern may show a scattered combination of nozzle blockages, while another could reveal a large number of neighboring nozzles failing simultaneously due to an electrical issue. 




To classify failures, manufacturers typically use two main approaches: rule-based methods and algorithmic-based models. Rule-based methods rely on organizing domain knowledge into a hierarchy of rules which produce classifications, while algorithmic approaches are data-driven, often leveraging \gls*{ml} techniques \cite{roehrich_predictive_2024}. Domain experts at \gls*{cpp} have developed an in-house rule-based classification model to identify failure mechanisms from the terminal nozzle log records based on expert knowledge and historical observations. Thus, when the classifier makes incorrect predictions for new data points, the rules need to be manually adjusted, or the predictions must be overwritten by hand. In contrast, \gls*{ml} models can automatically learn relationships between input data and output labels, allowing them to generalize better to unseen scenarios \cite{er_ratby_impact_2025}.



In this paper, we address a unique industrial challenge: classifying failure mechanisms of printheads at their \gls*{eol} stage using a multifaceted nozzle logging dataset. We propose an \gls*{ml}-based failure classification framework that integrates both time-series and spatial aspects of the data. Our approach builds on feature-based time-series classification methods, adapted to this specific application in collaboration with domain experts. Following extraction of useful features, an optimal classifier was built by fine-tuning and evaluating several traditional \gls*{ml} models and their \gls*{ovr} variants. Our results have shown that the \gls*{ovr} \gls*{rf} model performed best, achieving an average weighted F1 score of $0.93$ and reaching the human-level performance of the in-house rule-based baseline model (hereafter, the baseline).

\section{Related Work}\label{sec:background}

Industrial time-series classification datasets often focus on general manufacturing or industry-specific cases, such as automotive or healthcare. These datasets typically consist of conventional sensor readings capturing variables like vibration, temperature, and motion \cite{farahaniTimeseriesPatternRecognition2023}. A more complex case that happens in industry is in the time-series of historic log records such as sequences of error codes \cite{guillaumePredictiveMaintenanceEvent2021}.


In contrast, our problem resembles the identification of system failures based on the failure developments of its components. This results in a time-series representing the count information of how many component failures have been logged for the whole system. Thus, this type of data is sometimes formally referred to as the \textit{count} time-series \cite{davisCountTimeSeries2021} to highlight the difference between classical cases such as sensor-based. This problem is not well studied in the \gls*{ml} literature, but is mainly addressed with comprehensive statistical frameworks such as FMEDA \cite{grebe2007fmeda}, which relies on the estimates of failure rates of individual components. Moreover, existing research on CPP printheads has primarily addressed lifespan estimation \cite{parii2025predicting}, while classification of printhead failures from nozzle logging has not been studied yet. 

\gls*{tsc} has become a critical tool in modern industrial applications. It encompasses a variety of methods, among which feature-based algorithms are highlighted as an effective approach \cite{farahaniTimeseriesClassificationSmart2025}. Over the years, a wide range of feature-based techniques has been developed, from Shapelet-based like Fast Shapelets \cite{bostrom2015binary} to the hybrid state-of-the-art approaches such as HIVE-COTE 2.0 \cite{middlehurstHIVECOTE20New2021}. Another notable family of feature-based techniques focuses on statistical feature engineering. Recent frameworks like Tsfresh \cite{christ_time_2018} and TSFEL \cite{barandas2020tsfel} automate feature mining and selection using statistical tests, and they are particularly relevant for multi-variate time-series problems. However, these methods can be computationally intensive, as they often calculate a large number of features, many of which may be irrelevant to the specific problem \cite{tnaniEfficientFeatureLearning2022}. 

\section{Methodology}
\label{sec:concepts_and_approach}

\subsection{Nozzle logging and failure mechanisms of printers}

\begin{figure*}[h]
    \centering
    \includegraphics[width=1.0\textwidth]{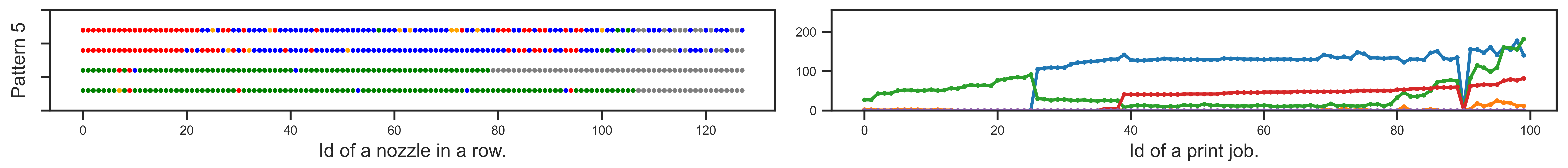}
    \caption{Example of a pattern in the nozzle log.}
    \label{fig:nozzle_grids}
\end{figure*}

\glsreset{nfc}


Typically, when printheads fail, it is possible to establish the associated failure mechanism based on the physical state or the error logging of that printhead. When several printheads fail for unknown reasons but exhibit similar nozzle log activity, it can be assumed they have the same failure mechanism. A printhead's nozzle log consists of several components among which one can find the status of individual nozzles according to \gls*{nfc}, which indicates that the nozzle does not jet the ink properly. If a nozzle is not behaving correctly, then it is assigned one of the five status labels \textit{Nozzle Failure [1-5]} (\textit{NF[1-5]}). The labels are anonymized to preserve the privacy of the internal terminology of \gls*{cpp}. 



Formally, a nozzle log of a printhead is a time series $\mathcal{T}^i$ which is defined as a sequence of log records:
\begin{equation}
    \mathcal{T}^i = \{X^{(1)}, X^{(2)}, \dots, X^{(n)}\},
    \label{eq:time_series}
\end{equation}
Where $n$ is the total number of log records and $i$ indexes the specific printhead. Each time series $\mathcal{T}^i$ of a removed printhead is associated with a corresponding failure mechanism label $y^i$. 
Each log record, $X^{(t)}$, captures the state of the printhead's 512 nozzles at a time-step $t$. The nozzles are arranged in four rows of length $128$ resulting in a matrix with dimensions $4 \times 128$. Each matrix entry indicates the state of a nozzle, where $NF[1-5]$ and $\emptyset$ denote possible \gls*{nfc} states:

\begin{multline}
    X^{(t)} \in \{NF1, NF2, NF3, NF4, NF5, \emptyset \}^{4 \times 128}, \\ \quad \forall t \in \{1, 2, \dots, n\}
    \label{eq:time_step_cat}
\end{multline}
Alternatively, each log record $X^{(t)}$ can be viewed as a 5-channel image of size $4 \times 128 \times 5$:
\begin{equation}
    X^{(t)} \in \{0,1\}^{4 \times 128 \times 5}, \quad \forall t \in \{1, 2, \dots, n\}.
    \label{eq:time_step_num}
\end{equation}
In this representation, $4$ and $128$ denote spatial dimensions of an image, corresponding to nozzle layout, while $5$ represents the number of channels. Each channel is a binary grid indicating whether a nozzle is in one of the five possible statuses $NF[1-5]$ at a time-step $t$. This representation allows to view nozzle log records as images, as illustrated in \cref{fig:nozzle_grids}, enabling visual identification of patterns. The grid plot on the left of \cref{fig:nozzle_grids} represents the full terminal record. The time-series on the right illustrates developments of the number of \glspl*{nfc} over time aligned over the last $100$ print jobs of a printhead. Each timestamp corresponds to the count of \glspl*{nfc} in the first nozzle log record of a print job while the color indicates the \gls*{nfc} type.

\subsection{Dataset}





The dataset used in this case study consists of $411$ printheads, which were removed from printers in the field due to failures in the nozzle log. The printheads in the dataset are categorized into six failure mechanisms of \textit{Pattern[1-5]} and \textit{Pattern 1\&2}. The class distribution in the dataset is heavily imbalanced, as the most populated class \textit{Pattern 1} accounts for $121$ samples, while the least populated class \textit{Pattern 5} has only $23$. Classes of \textit{Pattern 1} through \textit{Pattern 5} correspond to the failure mechanisms that can be clearly identified from the patterns in the nozzle log data. Cases that do not fit these patterns are grouped under a general \textit{Other} category, which includes printheads with sparse logs or atypical behaviors that lack sufficient data to constitute a separate class. The dataset also contains six printheads that exhibit characteristics of both \textit{Pattern 1} and \textit{Pattern 2}. While these instances could be treated as edge cases, we adopt a multi-label classification approach and include them as valid samples for both respective classes.

Class labels were assigned by the domain experts at \gls*{cpp} through a multi-stage process. Initially, failure mode predictions were generated using the in-house baseline model. Predicted labels were then verified by the domain experts, which involved a manual inspection of the terminal nozzle log record, usage history of each printhead and time-series information of \gls*{nfc} developments. Eventually, any incorrectly predicted labels were manually reviewed to ensure accuracy.

\subsection{Model design}

\begin{figure}[t]
    \centering
    \includegraphics[width=0.25\textwidth]{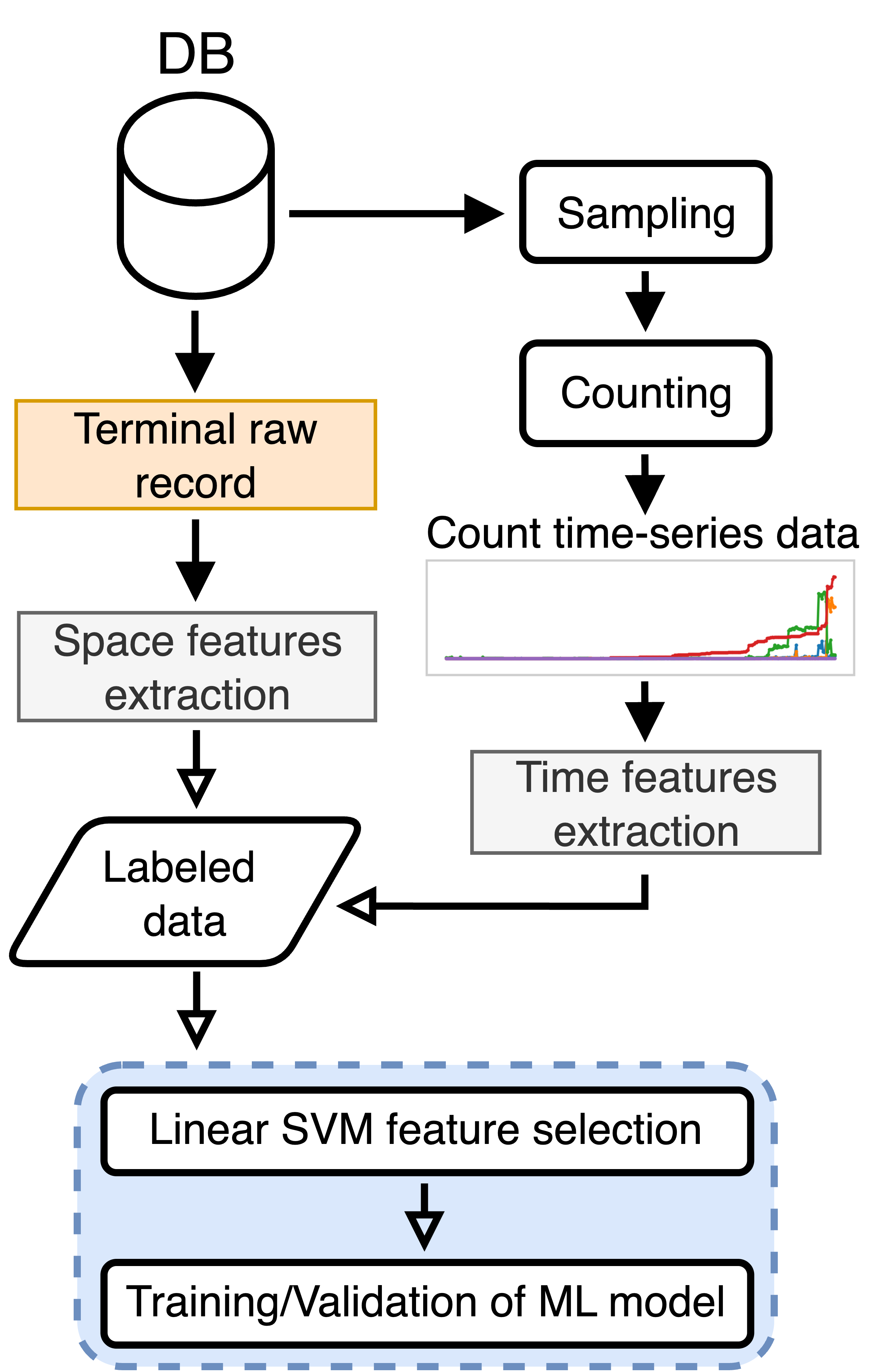}
    \caption{Overview of the modeling framework.}
    \label{fig:model_design}
\end{figure}


\subsubsection{Data extraction and representation}

A significant portion of the raw nozzle log may be superfluous, since new records are added with a high frequency, while \gls*{nfc} state is unlikely to change that rapidly. Additionally, raw nozzle log data poses challenges for storage and increases the time and space complexity of classification algorithms. Thus, when extracting raw nozzle log from the database (denoted as \textit{DB} in \cref{fig:model_design}), the data is sampled such that only the first record of every print job is considered. To obtain count time-series data from the nozzle log, the number of failed nozzles per \gls{nfc} type is taken from each time-step, transforming $\mathcal{T}^i$ into $\mathcal{T}'$. This results in a dataset of multi-variate time-series where each channel contains the count information of a specific nozzle failure type, as illustrated in \cref{fig:nozzle_grids}.

Next, we use Tsfresh Python package \cite{christ_time_2018} as it allows the extraction of various features from time-series. Out of the $60$ time-based feature functions provided by Tsfresh, only $11$ were selected based on a combination of domain expertise and empirical evaluation. The chosen features capture key characteristics such as linear trends, autocorrelation, complexity, and other relevant statistical properties. 

In addition to the Tsfresh-based features, seven custom functions were introduced to capture complementary aspects of the time series. These include the first and second derivatives, the final value of the time series, and the maximum difference between consecutive samples. The choice of these features was strongly guided by \gls*{cpp} domain experts. Derivative-based features enhance the model’s ability to distinguish subtle temporal variations, while the final value is critical as it plays a central role in the decision-making logic of the baseline model. 


Finally, space feature functions were added to extract properties from the terminal nozzle log state of printheads. Two such functions were added: the average position of failed nozzles per \gls*{nfc} and the number of consecutive NF4s from the edge of a grid. Overall, $20$ time-based feature functions were parameterized and applied to the nozzle log, resulting in $430$ numeric features per instance of a failed printhead.


\subsubsection{Development of an optimal classifier}
\label{sec:optimal_classifier}


Initially, an extensive set of \gls*{ml} models available through the \textit{scikit-learn} library was evaluated with the \textit{pycaret} package. Based on evaluation results, it was decided to proceed with the \gls*{rf}, \gls*{lr}, \gls*{et}, \gls*{dt} and \gls*{knn} methods, as well as with \gls*{svm} with a \gls*{rbf} kernel. For each of these methods, a pipeline of imputation, standard scaling and feature selection was used. To discard irrelevant features and improve generalization, a model-based feature selection procedure was employed using a linear kernel \gls*{svm} with a tight regularization coefficient of $0.01$. These components of the modeling framework are represented by the blue rectangle in \cref{fig:model_design}.

Finally, the problem was transformed into a multi-label classification setting, as a handful of samples exhibited both \textit{Pattern 1} and  \textit{Pattern 2} failure mechanisms at the end of their lifetime. To allow predicting multiple labels, classifiers were adapted to the \gls{ovr} framework. Specifically, a separate binary classifier is trained for each class against all others, and the final decision is obtained by combining their outputs. 

\section{Evaluation}
\label{sec:results}


\subsection{Evaluation procedure and metrics}


Due to the imbalanced dataset and a limited number of samples for several classes, all of the comparisons between the models and training settings were performed using the \gls*{loocv} framework. After the split-wise predictions were obtained, the precision, recall and F1 scores were subsequently calculated. Precision can be interpreted as the ratio of correctly predicted samples over the total number of predictions, while recall corresponds to the ratio of true samples that were accurately classified. The F1 score aggregates these two values with a harmonic mean, providing an overall outlook on the performance of the model. For the evaluation of multi-labeled classifiers, it is necessary to properly average out the scores according to one of the strategies, namely micro-average, macro-average, and weighted-average. We used weighted-average as this type of averaging is particularly suitable for classification problems with class imbalance as it ensures a fairer evaluation.




\subsection{Results}

\subsubsection{Selection of the highest performing ML model}

To identify the best performing \gls*{ml} model, the hyperparameters for each classifier were tuned using a $10$-fold cross validation procedure. Next, the \gls*{loocv} performance was assessed for each of the models with the weighted scores being summarised in \cref{tab:models_scores}. 
Since the primary focus of this experiment is on discriminating between the five well-defined failure patterns, the weighted averages were computed by excluding the \textit{Other} class from the evaluation.
As it can be observed from the \cref{tab:models_scores} the \acrlong*{rf} consistently outperforms other classifiers in terms of the F1 score and recall. Based on these findings, a \acrlong*{rf} model with $50$ trees, a maximum depth of $20$ and the Gini impurity criterion was selected for use in the subsequent round of experiments.

\begin{table}[t]
    \centering
    \begin{tabular}{lccc}
    \toprule
    \textbf{Model} & \textbf{Prec.} & \textbf{Rec.} & \textbf{F1} \\
    \midrule
    \gls*{rf}  & 0.9318 & 0.9513 & 0.9410 \\
    \gls*{lr}  & 0.9201 & 0.9294 & 0.9242 \\
    \gls*{knn} & 0.8366 & 0.8394 & 0.8369 \\
    \gls*{dt}  & 0.8077 & 0.8881 & 0.8428 \\
    \gls*{et}  & 0.9355 & 0.9440 & 0.9392 \\
    \gls*{svm} & 0.9170 & 0.9367 & 0.9266 \\
    \bottomrule
    \end{tabular}
    \caption{Evaluation results of the tuned ML models.}
    \label{tab:models_scores}
\end{table}

\subsubsection{Evaluation of an optimal ML classifier against the rule-based baseline}

Following the identification of the optimal classification model in the initial experiment, it is now possible to compare the performance of the developed classifier to the baseline on the full dataset. While direct comparisons between rule-based and \gls*{ml} methods are typically inappropriate due to their fundamentally different algorithmic approaches, such a comparison is justified in this case. The rule-based baseline, developed by domain experts, establishes a higher validation boundary than a random classifier. Furthermore, the absence of \gls*{ml} baselines for this specific problem makes the rule-based approach a reasonable benchmark against which our model's performance can be evaluated.

The evaluation scores summarizing the performance of the baseline and \gls*{ml} model are presented in \cref{tab:classification_report}. From the results, it is evident that the \gls*{ml} model clearly outperforms the baseline in predicting \textit{Pattern 2}, \textit{Pattern 4}, and \textit{Pattern 5} as reflected in terms of the F1 scores. These findings are further supported by the confusion matrices in \cref{fig:confusion_results}. Note that because OVR classifiers may output multiple predictions, the raw sums can exceed the actual class support values. However, this does not undermine their usefulness for comparing class-wise prediction tendencies. At the same time, the baseline demonstrates strong superiority in predicting \textit{Pattern 3} failures and performs slightly better for \textit{Pattern 1}, while the \gls*{ml} model almost reaches its recall at $0.96$. 
For \textit{Pattern 2}, the \gls*{ml} model has a higher precision of $0.95$, however the baseline outperforms its recall with an almost perfect score of $0.99$. Finally, in terms of overall misclassifications, the \gls*{ml} model demonstrates improved performance, with $31$ incorrect predictions compared to $39$ from the baseline.

\begin{table}[t]
    \centering
    \begin{tabular}{llcccc}
    \toprule
     &Model& \textbf{Prec.} & \textbf{Rec.} & \textbf{F1} & \textbf{Support} \\
    \midrule
    Pattern 1  &Baseline& 0.93 & 0.98& 0.95 & 127\\
    & OVR RF& 0.90& 0.96& 0.93&\\
    \midrule
    Pattern 2  &Baseline& 0.80 & 0.99 & 0.89 & 75 \\
    & OVR RF& 0.95& 0.93& 0.94&\\
    \midrule
    Pattern 3  &Baseline& 1.0 & 1.0 & 1.0 & 30 \\
 & OVR RF& 0.80& 0.93& 0.86&\\
 \midrule
     Pattern 4  &Baseline& 0.76 & 0.73 & 0.75 & 26 \\
 & OVR RF& 1.00& 0.85& 0.92&\\
 \midrule
    Pattern 5  &Baseline& 0.92 & 0.52 & 0.67 & 23 \\
 & OVR RF& 0.95& 0.87& 0.91&\\
 \midrule
    Other  &Baseline& 0.97& 1.00& 0.99& 136\\
 & OVR RF& 0.96& 0.93& 0.94&\\
 \midrule
    Weighted  &Baseline& 0.91& 0.95& 0.93& 417\\
 average & OVR RF& 0.93& 0.93& 0.93&\\
 \bottomrule
    \end{tabular}
    \caption{Classification report of the rule-based baseline and OVR Random Forest model for the five common output labels, transformed into multi-label problem.}
    \label{tab:classification_report}
\end{table}

\begin{figure}[ht]
\footnotesize
\centering
\begin{tikzpicture}[scale=0.7]
    \node at (3, 0) {\small Baseline};

    \foreach \y [count=\n] in {
{ 124, 12, 0, 0, 1, 0},
{ 2, 74, 0, 0, 0, 0},
{ 0, 0, 30, 0, 0, 0},
{ 6, 4, 0, 19, 0, 1},
{ 2, 2, 0, 6, 12, 3},
{ 0, 0, 0, 0, 0, 136},
    } {
        \foreach \x [count=\m] in \y {
            \definecolor{customblue}{RGB}{255,178,178}
            \pgfmathsetmacro\scaledX{100*\x/100} 
            \pgfmathsetmacro\opacityValue{\scaledX/100} 
            \node[fill=customblue!\scaledX!white, minimum size=8mm, text=black] at (\m,-\n) {\x};
        }
    }

    \foreach \a [count=\i] in {Pattern 1, Pattern 2, Pattern 3, Pattern 4, Pattern 5, Other} {
        \node[minimum size=1mm, rotate=45, anchor=north east] at (\i, -6.5) {\a};
    }
    \foreach \a [count=\i] in {Pattern 1, Pattern 2, Pattern 3, Pattern 4, Pattern 5, Other} {
        \node[minimum size=1mm] at (-0.5,-\i) {\a};
    }
    \node[rotate=90] at (-1.8, -3.5) {\small True Class};
    \node at (3.5, -8.3) {\small Predicted Class};
    
\end{tikzpicture}
\vfill
\begin{tikzpicture}[scale=0.7]
        \node at (3, 0) {\small OVR RF};

    \foreach \y [count=\n] in {
{ 122, 3, 1, 0, 0, 2},
{ 3, 70, 0, 0, 0, 0},
{ 1, 0, 28, 0, 0, 3},
{ 4, 1, 0, 22, 0, 0},
{ 3, 0, 0, 0, 20, 0},
{ 3, 0, 6, 0, 1, 126},
    } {
        \foreach \x [count=\m] in \y {
            \definecolor{customblue}{RGB}{255,178,178}
            \pgfmathsetmacro\scaledX{100*\x/100} 
            \pgfmathsetmacro\opacityValue{\scaledX/100} 
            \node[fill=customblue!\scaledX!white, minimum size=8mm, text=black] at (\m,-\n) {\x};
        }
    }

    \foreach \a [count=\i] in {Pattern 1, Pattern 2, Pattern 3, Pattern 4, Pattern 5, Other} {
        \node[minimum size=1mm, rotate=45, anchor=north east] at (\i, -6.5) {\a};
    }
    \foreach \a [count=\i] in {Pattern 1, Pattern 2, Pattern 3, Pattern 4, Pattern 5, Other} {
        \node[minimum size=1mm] at (-0.5,-\i) {\a};
    }
    \node[rotate=90] at (-1.8, -3.5) {\small True Class};
    \node at (3.5, -8.3) {\small Predicted Class};
    
\end{tikzpicture}
\caption{Confusion matrices of prediction results of the rule-based baseline model and proposed ML model across common output labels.}
\label{fig:confusion_results}
\end{figure}

\subsubsection{Feature importance in class predictions}

The use of a \acrlong*{rf} model as the primary classification approach allows for clear interpretation of feature importance through the Gini index. By analyzing the top $10$ features for each class, we gained insights into the model's decision-making process. Notably, custom features created with domain knowledge, such as the number of consecutive NF4s, maximum differences and derivatives, were among the most relevant for predicting \textit{Pattern 2}. Similarly, for \textit{Pattern 1}, maximum differences and linear trend features derived from domain expertise were given the highest importance scores. 
In addition, domain-informed features also played a significant role for the correct classification of \textit{Pattern 4}.

\subsection{Discussion}

The results from the first experiment indicate that the \acrlong*{rf} model demonstrates superiority in terms of F1 score and recall, likely due to its ability to learn non-linear decision boundaries through ensemble learning. The \gls*{lr} and \gls*{svm} models perform similarly well, possibly benefiting from the well-designed set of features. In contrast, \gls*{dt} and \gls*{knn} exhibit the worst performance, likely due to overfitting and poor generalization on the small dataset.

In the main experiment of this study, the shortcomings of the rule-based model were clearly summarized in the classification report for its predictions in \cref{tab:classification_report} and in the confusion matrices on \cref{fig:confusion_results}. The performance drawbacks mainly stem from the high number of \glspl*{fp} for \textit{Pattern 2} and the high number of \glspl*{fn} for \textit{Pattern 5}. This indicates that the baseline algorithm has a simple criterion for assigning \textit{Pattern 2} and struggles to detect many instances of \textit{Pattern 5}. Additionally, a relatively high number of \glspl*{fn} is present for \textit{Pattern 4}.

It was demonstrated that an \gls*{ml} classifier outperformed the baseline in terms of the number of misclassified samples and achieved a comparable performance in terms of the weighted F1 score. Key limitations include struggles with the class \textit{Other}, as the $10$ instances of this class were misclassified. Additional confusions occurred between \textit{Pattern 1} and \textit{Pattern 2} classes, which even domain experts find hard to distinguish. Errors also arose for the \textit{Pattern 3} samples where nozzle log was mostly empty and only several failure records were present at the terminal stage of the printhead's lifespan, suggesting that time-based feature functions could not adequately capture such edge cases. This highlights the need for further refinement of features, though some cases remain challenging to discriminate even for domain experts.

The feature importance scores retrieved for the Random Forest model provided a valuable insight that most of the custom features designed with the domain knowledge were important for predicting the \textit{Pattern 1}, \textit{Pattern 2} and \textit{Pattern 4} classes. 


\section{Conclusion}

This study has addressed the problem of classification of failure mechanisms of \gls*{cpp}'s printheads caused by errors in nozzle logging highlighting the importance of synergy between data-driven approaches and domain knowledge in industrial corrective maintenance. An \gls*{ml} classifier was successfully developed and demonstrated to outperform the rule-based baseline for specific failure patterns, reaching an overall F1 score of $0.93$. It was determined that \gls*{ovr} \gls*{rf} performed best across evaluated models and exceeded the predictive performance of rule-based baseline with respect to three out of five classes. It was determined that the problem of multi-variate count time-series classification in the context of limited data can be effectively addressed by transforming each individual time-series into a fixed-length feature vector using time-based feature functions. The proposed classification framework and feature engineering approach are relevant for analyzing similar systems of multiple small parts prone to failure or degradation over time.
 




\section*{Acknowledgment}
We would like to thank \gls*{cpp} for providing access to data
used for model development and evaluation in this work as well as their guidance and support throughout the research and implementation process.


\end{document}